\documentclass[runningheads]{llncs}
\usepackage[T1]{fontenc}
\usepackage{graphicx, amsmath}
\usepackage{wrapfig}
\usepackage{tikz}
\usepackage{subcaption}
\usepackage{url}
\usetikzlibrary{shapes.geometric, arrows.meta, positioning, calc, backgrounds}
\usepackage{booktabs}
\usepackage{threeparttable}

\begin{document}
%

%
\title{Imitation learning for clinical decision support in pediatric ECMO\thanks{We acknowledge support from NIH award R01NS133142}}
\titlerunning{Imitation learning for pediatric ECMO}

\author{Fateme Golivand Darvishvand\inst{1} \thanks{Corresponding author, \email{fateme.golivanddarvishvand@utdallas.edu}}\and
Michael Skinner\inst{1} \and
Saurabh Mathur\inst{2}\and
Ameet Soni\inst{3} \and
Phillip Reeder\inst{4} \and
Kristian Kersting\inst{2}\and
Lakshmi Raman\inst{4}\and
Sriraam Natarajan\inst{1}}
\authorrunning{Golivand et al.}

\institute{University of Texas at Dallas \and
Technische Universität Darmstadt
\and
Swarthmore College\and
University of Texas Southwestern Medical Center}
%
\maketitle              
\begin{abstract}
Pediatric critical care is a dynamic, high-stakes process involving constant monitoring and adjustments in life-saving treatments. Modeling these interventions is crucial for effective decision support. To address the challenges of high complexity and data scarcity in pediatric Extracorporeal Membrane Oxygenation (ECMO), we frame clinical decision-making as learning to act from trajectories, i.e., imitation learning that learns action models from observational data, with a key feature that actions are not directly observed. We consider TabPFN, a recent transformer-based approach for tabular data, and traditional baselines including XGBoost and Multi-Layer Perceptrons (MLPs) on real-world pediatric ECMO data to learn the action models. We find that the TabPFN-based approach consistently outperforms these classical baselines, supporting its use as a strong clinician-behavior baseline for pediatric ECMO decision support.

\end{abstract}
\section{Introduction}

Generative AI has made significant progress in the past few years and has been widely adopted for prediction tasks~\cite{sengar2025generative,vaswani2017attention}. While previously it was acknowledged that these methods do not handle tabular data such as EHRs well, recent work in this direction has made crucial progress~\cite{hollmann2025accurate}. However, their use in modeling policies, such as choosing a sequence of actions, remains relatively unexplored. In high-stakes settings such as ICUs, there is a need to make decisions sequentially and continuously.

We explore the problem of ``learning to act'' by observing physicians in a pediatric Extracorporeal Membrane Oxygenation (ECMO) unit. This problem is particularly challenging for several reasons -- (1) The time between actions is irregular; (2) The action space is concurrent as several parameters could be changed simultaneously; (3) A large action space; (4) A large number of observations (actions) for a small number of patients, i.e., imbalance of the features vs examples; (5) Finally, there is no explicit reward signal and one has to infer the success of a trajectory based on the patient outcome. 

To address these challenges, we model the learning to act problem as ``imitation learning'' where the AI agent observes a sequence of events and learns a policy (a mapping from observations to actions). While imitation learning has been extensively researched~\cite{hussein2017imitation,natarajan2011imitation}, it has not been considered for learning to act in pediatric ECMO treatment or similar contexts, posing the challenges outlined above. We consider three different types of learners -- multilayer perceptron, gradient-boosting, and the more recent tabular foundation model. We evaluate these models as Imitation Learning baselines for sequential clinical decision support from offline trajectories, and empirically find the tabular foundation model to be a particularly strong clinician-behavior model in this setting.
Our empirical evaluation demonstrates that imitation learning holds promise for this challenging data as a tool for clinical decision support\footnote{Code is available at \url{https://github.com/fateme-gd/ImitationLearning_ECMO.git}}.

The rest of the paper is organized as follows: after presenting the necessary background on the problem and the data, we outline our key contribution of imitation learning when learning to act in pediatric ECMO. We present details of the proposed framework before performing qualitative and quantitative evaluations on ECMO data. Finally, we conclude by discussing the findings and outlining areas of future research.

\vspace{-0.5em}
\section{Extracorporeal Membrane Oxygenation (ECMO)}

Extracorporeal Membrane Oxygenation (ECMO)~\cite{ecmo} is a method for supporting severely ill patients with cardiac or respiratory failure. Managing an ECMO circuit requires clinicians to make continuous, high-stakes adjustments to machine parameters like gas flow and oxygen concentration. Moreover, these decisions must be made based on rapidly shifting physiological signals. Despite the critical nature of these decisions, there is a lack of automated decision support tools for this domain, especially for pediatric patients.

Developing effective decision support in this domain is challenging due to several factors in addition to the ones mentioned above. First, the lack of complete and formalized knowledge about pediatric ECMO limits the efficacy of model construction. Second, gaps in domain knowledge cannot be filled through the gold standard of hypothesis testing since evaluating new strategies directly on live patients is ethically and practically unfeasible in a critical care setting. In the next section, we present our approach to take the first step in building a decision support system for this challenging task.

\section{Learning to act in Pediatric ECMO}

We now present our framework for learning treatment policies by identifying patterns in expert behavior from historical data. At a high level, we formalize this task as follows:


\begin{center}
\fbox{%
  \begin{minipage}{\dimexpr\linewidth-2\fboxsep-2\fboxrule\relax}
  \raggedright
  \textbf{Given:} An observational dataset of pediatric ECMO patient trajectories.\\[2pt] 
  \textbf{Task:} Obtain a distribution over a treatment plan (i.e., a policy) that imitates clinician behavior.

  \end{minipage}%
}
\end{center}

We solve the problem using the framework of Imitation Learning (IL)~\cite{hussein2017imitation}. Unlike standard reinforcement learning~\cite{sutton2018reinforcement}, which requires active environmental interaction, a process that is ethically and practically impossible in a pediatric ICU, offline IL allows us to exploit expert demonstrations already present in historical clinical records.

\begin{figure}[t!]
\vspace{-.3cm}
    \centering
    \includegraphics[width=0.75\textwidth]{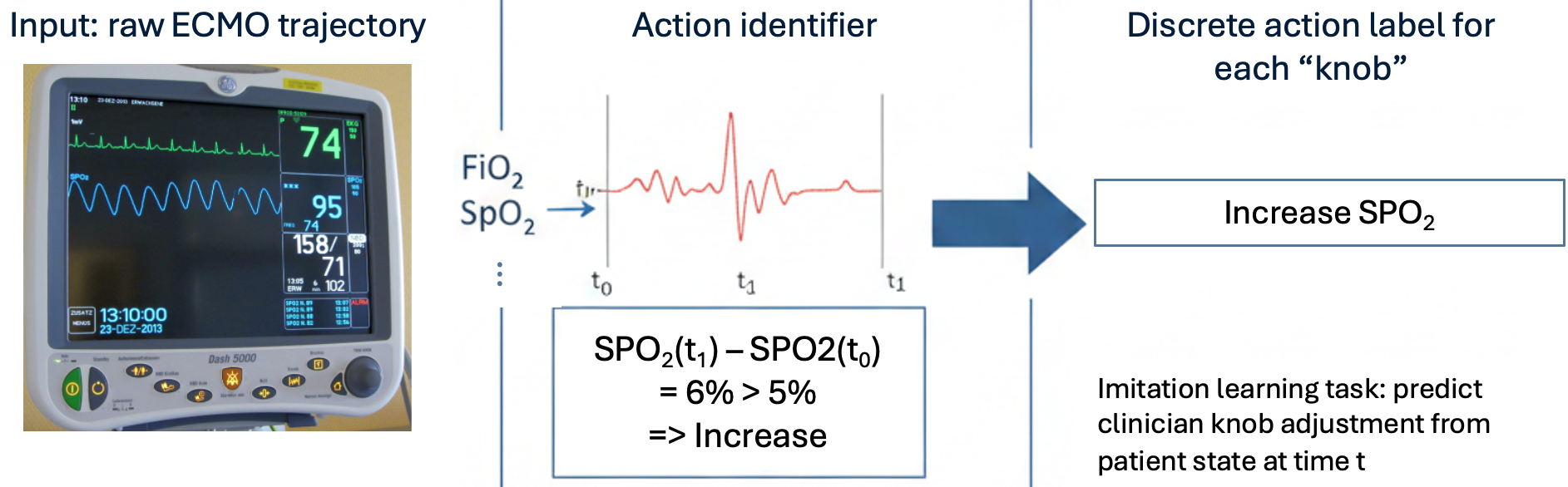}
    \caption{\small Pipeline for Learning Clinical Behavior from Unlabeled ECMO Trajectories. (Left) Raw physiologic telemetry is processed into discrete action labels using physician-defined thresholds (e.g., $\Delta SpO_2>5\%$). (Right) These discovered actions formulate an imitation learning task: predicting clinician ``knob'' adjustments from patient state.}
    \label{fig:setup_pipeline}
\end{figure}

We first formalize the problem by defining the trajectory as a sequence of {\it patient states} and {\it clinician actions}. A state is, formally, a snapshot of the world -- the current observations on the ECMO machine, i.e., the values of the parameters that are being monitored along with the general state of the patient. 
As mentioned earlier, interventions are not explicitly timestamped in the raw dataset. Hence, it is necessary to infer them from observations. As illustrated in Figure~\ref{fig:setup_pipeline}, we define actions in terms of special control variables as a change exceeding a physician-defined threshold. For example, if we find that over a time interval the patient exhibits an increase in arterial P$\text{O}_2$ from 50mmHg to 100mmHg, we infer that the clinician has acted to effect this observed physiologic change. Finally, we consider three different methods for learning action models. In addition to the classical methods, we consider a new transformer-based Prior-Data Fitted Network (TabPFN~\cite{hollmann2025accurate,grinsztajn2025tabpfn,hollmann2023tabpfn}) method.

\subsection{Defining States and Actions}
We formalize the observational trajectories as a sequence of tuples $(s_t,a_t).$ Modeling of pediatric ECMO interventions requires a high-dimensional representation of both the patient's condition (the state) and the clinician's response (the action). Physicians monitor multiple clinical features of the patient, providing critical indicators of their health. These physiological features include hemodynamics, gas exchange metrics, and laboratory test results. These values describe the overall physiological state at time $t$ and are represented as
$s_t = (s_t^1,\dots,s_t^N)$.

In addition to monitoring, physicians adjust several treatment components to maintain stability. We refer to these as {\it actionable features}, detailed in Table \ref{tab:action_scales}. As a result, the intervention at time $t$ is multi-dimensional $a_t = (a_t^1,\dots,a_t^M)$, where each $a_t^k$ corresponds to an adjustment on one specific component (a ``knob'').

Since raw clinical records do not contain explicit labels for these multi-dimensional interventions, we derive them using a heuristic grounded in clinical expertise.
As shown in Figure~\ref{fig:setup_pipeline}, for each of the four actionable features (knobs), we derive discrete action labels by monitoring the delta over a fixed aggregation window. For a feature $x$ at time $t$, the label $a_t^k$ is:
\[a_t^k =
\begin{cases}
\text{Increase} & \text{if } (x_{t+1}^k - x_t^k) > \delta_k \\
\text{Decrease} & \text{if } (x_{t+1}^k - x_t^k) < -\delta_k \\
\text{Same} & \text{otherwise}
\end{cases}\]
where $\delta_k$ is the physician-defined threshold (e.g., 25 mmHg for PO2) indicating a change in setting. This ensures the labels distinguish significant clinical interventions from sensor noise and physiological fluctuation. Having formalized the trajectories as temporal sequences of state-action tuples, $(s_t,a_t)$, we now describe our learning procedure.

\subsection{Imitation Learning}
\begin{wraptable}[15]{r}{5.5cm}
\vspace{-.8cm}
\centering\small
\caption{Actionable features (knobs) and physician-defined action thresholds. A discrete action is identified when the absolute change within a 60-minute window exceeds the specified threshold.}
\label{tab:action_scales}
\begin{tabular}{lr}
\toprule
\textbf{Knob} & \textbf{Threshold} \\
\midrule
Arterial PO$_2$   & 25 mmHg \\
Arterial PCO$_2$  & 5 mmHg \\
SpO$_2$           & 5  \% \\
FiO$_2$     & 10 \% \\
\bottomrule
\end{tabular}
\end{wraptable}
Imitation learning aims to learn a policy by mimicking expert behavior. It learns expert behavior from demonstrations in the form of a dataset of trajectories. 
In general, expert actions at each time-step $t$ may depend on the full history of states. Let us denote the distribution over expert actions as $p(A_t \mid s_{1:t}),$ where $A_t$ is a random variable representing the action at time $t$ and $s_{1:t}$ is the sequence of clinical observations up to time $t$. However, modeling the full history directly is computationally intractable and can lead to overfitting, especially in data-scarce domains. 

To make the learning problem tractable, we make the first-order Markov assumption that the current state representation $s_t$ is a sufficient statistic of the relevant history~\cite{puterman2014markov}. This allows us to simplify our model of the expert's decision rule to $p(A_t \mid s_t)$, effectively transforming the sequential decision-making problem into a sequential supervised learning task.

\begin{figure}[ht!]
    \centering
    \resizebox{.6\textwidth}{!}{
\begin{tikzpicture}[
    node distance=1.2cm and 2cm,
    block/.style={draw, thick, minimum width=3cm, minimum height=0.8cm, font=\sffamily},
    decision/.style={draw, thick, fill=blue!5, minimum width=3.5cm, minimum height=1cm, font=\sffamily\bfseries},
    outcome/.style={draw, dashed, fill=gray!5, minimum width=2.5cm, font=\sffamily\small},
    arrow/.style={-Stealth, thick}
]

    \node [block] (head) {Knob Head $k$};

    \node [decision, below=of head] (change) {Change Indicator $c_t^k$};
    
    \node [decision, below right=1.5cm and 0.5cm of change] (dir) {Direction $d_t^k$};

    \node [outcome, below left=1.5cm and 0.5cm of change] (same) {$y_t^k =$ Same};
    \node [outcome, below=1cm of dir, xshift=-2.4cm] (dec) {$y_t^k = $ Decrease};
    \node [outcome, below=1cm of dir, xshift=2.4cm] (inc) {$y_t^k =$ Increase};

    \draw [arrow] (head) -- (change);
    
    \draw [arrow] (change) -| node[pos=0.3, above] {No (0)} (same);
    \draw [arrow] (change) -| node[pos=0.3, above] {Yes (1)} (dir);
    
    \draw [arrow] (dir) -| (dec);
    \draw [arrow] (dir) -| (inc);

\end{tikzpicture}
    
    }
    \caption{Multi-head policy architecture for simultaneous multi-knob control. The model processes the system state to output a joint action vector $a_t$ across four independent knobs. For each knob k, the policy predicts a 3-class action $y_t^k$ through a hierarchical logic: first determining if a change is required ($c_t^k$), followed by the direction of change ($d_t^k$). This independent multi-head approach avoids the dimensionality explosion of modeling the full joint action space.}
    \label{fig:label_hierarchy}
\end{figure}
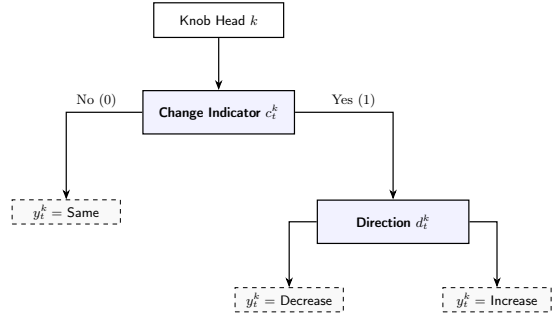
Moreover, recall that since clinicians may adjust several machine parameters simultaneously, $A_t$ is a composite action, defining the change over $M$ knobs. Since each knob $k \in \{1,\dots,M\}$ may be increased, decreased, or kept the same, $A_t$ can have $3^M$ possible values, making the learning problem challenging. To address this, we model each ``knob'' $k$ using an independent predictor $p_k(A_t^k \mid s_t)$. This multi-head approach allows for simultaneous adjustments while avoiding the combinatorial explosion of modeling the full joint action space.

Finally, since clinicians do not intervene on every knob at every time step, most knobs are kept the same, making the classes imbalanced. We address this by employing a hierarchical two-stage prediction pipeline. First, a binary classifier predicts the change indicator $c_t^k$. At inference, the model produces a probability that a change occurs given state $s_t$, $p(c_t^k=1\mid s_t)$, which is thresholded by a value $\tau_k$ optimized on the training data. Next, if the change indicator is true, a second binary classifier determines the direction of this change $d_t^k\in\{\text{Decrease},\text{Increase}\}$. This model is trained only on samples where an intervention occurred ($c_t^k=1$) to focus on the physiological triggers for adjustments. 

The final output for each knob is merged into a 3-class action prediction 
${y}_t^k$\vspace{-0.5em}
\[{y}_t^k =
\begin{cases}
\text{ Same} & \text{if } {c}_t^k = 0 \\
\text{ Decrease} & \text{if } {c}_t^k = 1 \text{ and } {d}_t^k = \text{Decrease} \\
\text{ Increase} & \text{if } {c}_t^k = 1 \text{ and } {d}_t^k = \text{Increase}
\end{cases}\]
The overall multi-knob intervention at time $t$ is represented by the vector $a_t = (y_t^1,\dots,y_t^M).$ Figure \ref{fig:label_hierarchy} summarizes this hierarchical setup.
For learning the predictors, we consider 3 methods: (i) a multi-layer perceptron (MLP) that directly learns the actions, (ii) gradient-boosted decision trees (XGBoost) that learns a distribution over the actions,  and (iii) TabPFN, which uses a transformer-based prior-data fitted network that uses in-context learning to directly output posterior predictive probabilities without gradient-based training. Our goal in using multiple methods is not only to ascertain the usefulness of these methods but also to understand if the choice of the model matters when learning to imitate the expert. We now briefly describe each method.

\begin{itemize}\itemsep0em 
\item {\bf Multi-Layer Perceptron (MLP):}
MLP~\cite{goodfellow2016deep} is a deep feedforward neural network that takes an input feature vector (here the patient's clinical state), passes it through multiple fully connected layers with nonlinear activations, and produces logits $\psi_\theta(a,s)$ (clinical interventions) that are turned into probabilities using a softmax. MLPs are trained by minimizing a loss, using backpropagation and gradient-based optimization.
\item{\bf XGBoost:}
XGBoost~\cite{chen2016xgboost} learns using gradient boosted decision trees, building the model additively as an ensemble of weak learners (shallow decision trees) trained sequentially to correct prior errors. XGBoost is a strong tabular baseline because tree splits capture nonlinearities and the algorithm implements many strategies to prevent overfitting to the data, such as early stopping, weight penalties, and pruning.  It has been shown to be effective on a wide variety of machine learning tasks and data sets.

\item{\bf TabPFN:}
TabPFN~\cite{hollmann2025accurate,grinsztajn2025tabpfn,hollmann2023tabpfn} is a \emph{prior-data fitted network} for tabular prediction. It is a transformer pretrained on many synthetic tasks sampled from a broad prior. This pretraining allows TabPFN to perform probabilistic inference on new tabular datasets by conditioning on the labeled examples provided at test time. It does that with a single forward pass of the dataset and does not rely on iterative gradient-based training and hyperparameter tuning for that dataset. Given a set of training pairs $\{(s_n,a_n)_{n=1}^{N}\}$ and a query state $s$, TabPFN outputs posterior predictive probabilities $p(a\mid s)$ in a single forward pass, making it an interesting model for IL where we want an estimate for $p(a\mid s)$ from limited trajectories.

\end{itemize}


\section{Evaluation}

We aim to answer the following research questions:

\begin{itemize}
    \item\textbf{Q1}: Do the learned policies generalize to unseen data?
    \item\textbf{Q2}: Are the policies
    calibrated? 
    \item\textbf{Q3}: Where do the learned policies disagree with the clinicians?
\end{itemize}

\subsection{Data Set Creation}

To answer these research questions, we utilized a dataset of 78 pediatric ECMO trajectories obtained from the Children's Medical Center, Dallas. This dataset excludes cases with Congenital Heart Disease since they require specialized management owing to higher complexity. The average age of subjects was 4.66 years old, and the average length of their trajectory was 215 hours. We discretized the observations into hourly windows and aggregated values by mean.

Each trajectory consisted of 4 kinds of data: hemodynamics, ECMO circuit configuration, ventilator readings, and laboratory test results. From these, we focus on 23 physician-selected variables. We used these variables to define the patient state at a particular hour $t$ as follows. For each variable, we included its value at hour $t$ and the difference relative to the previous hour. Age-sensitive variables heart rate (HR) and mean arterial pressure (ARTm) were normalized according to the patient's age. Finally, we also included the type of ECMO (VV or VA), and a binary indicator variable (\textit{on-ecmo}) denoting whether the patient is on ECMO at a given time step. Overall, the state at each hour consists of 48 values: the current value and previous-hour delta for each of the 23 physiologic variables, plus ECMO type and the binary \textit{on-ecmo} indicator. Table~\ref{tab:state_features} summarizes the 23 physiologic variables.

\begin{table}[t]
\vspace{-.5cm}
\centering
\caption{Physiologically curated state features used to construct $s_t$. For each feature, we include the mean value within a 60-minute window and its delta relative to the previous window. HR and ARTm are age-normalized.}
\label{tab:state_features}
\small 
\resizebox{.75\textwidth}{!}{
\begin{tabular}{lll}
\toprule
\textbf{Category} & \textbf{Feature Name} & \textbf{Unit} \\
\midrule
\textbf{Hemodynamics} & Mean Arterial Pressure (ARTm) & mmHg \\
& Heart Rate (HR) & bpm \\
& SpO2 & \% \\
& Cerebral Oximetry (rSO2-1, rSO2-2) & \% \\
\midrule
\textbf{ECMO Circuit} & Blood Flow & mL/kg/min \\
& Sweep Gas: CO2 Flow & L/min \\
& Sweep Gas: O2 Flow & L/min \\
& FiO\textsubscript{2} -- ECMO& \% \\
& Volume Sensor & mmHg \\
\midrule
\textbf{Ventilator} & Mean Airway Pressure ($P_{AW}$) & cmH$_2$O \\
& Positive End-Expiratory Pressure (PEEP) & cmH$_2$O \\
& Oxygen Concentration (FiO2) & \% \\
& Tidal Volume & mL/kg \\
& End-tidal carbon dioxide (etCO2) & mmHg \\
\midrule
\textbf{Laboratory} & Arterial partial pressure of oxygen (PO2)  & mmHg \\
& Arterial partial pressure of carbon dioxide (PCO2)  & mmHg \\
& Arterial pH & - \\
& Arterial Base Excess & mmol/L \\
& Lactate & mmol/L \\
& Ionized Calcium & mmol/L \\
& Total CO2 & mEq/L \\
& International Normalized Ratio (INR) & - \\
\bottomrule
\end{tabular}

}
\end{table}


Since the raw dataset does not contain explicit intervention labels, we construct discrete action labels using physician-provided thresholds for 4 frequently changing actionable variables. For each such variable, we mark a ``change'' when the observed delta over one time slot exceeds the given threshold, as discussed in Section 3. Table ~\ref{tab:action_scales} summarizes these 4 actions and their thresholds.  

\vspace{-1em}
\subsection{Evaluation protocols and baselines}

We consider three approaches for predicting actions given the state of a patient: XGBoost, multi-layer perceptron (MLP), and TabPFN. XGBoost models $\psi_\theta$ using gradient-boosted decision trees, learning an additive ensemble of 200 shallow trees. The MLP consists of a feedforward neural network of 3 fully connected layers and nonlinear activations, trained end-to-end with gradient-based optimization. For TabPFN, we utilized version 2.5. We evaluate these approaches using leave-one-out (LOO) cross-validation:
in each fold, one patient trajectory is held out for testing, and all remaining patients are used for training. We apply the same LOO protocol and an identical two-stage pipeline for all three methods.


\subsubsection{Metrics}

To evaluate our models, we consider metrics that are appropriate for imbalanced data sets (most actions in the trajectories are ``same'').  We report balanced accuracy and macro-F1 for each action. We additionally report overall macro-F1, macro-precision, and macro-recall by aggregating performance across all actions. Accuracy alone can be misleading in this setting, since a model that over-predicts the majority ``same'' class may appear to perform well while failing to detect clinically relevant actions. In contrast, balanced accuracy and macro-averaged metrics weight classes more evenly, providing a clearer view of performance on minority actions (change vs no-change). Finally, we report Expected Calibration Error (ECE)~\cite{naeini2015obtaining} to assess overall calibration performance. ECE summarizes the average gap between a model's stated confidence and its observed accuracy across probability bins, so lower ECE indicates that predicted probabilities can be interpreted as more reliable estimates of uncertainty.
\subsection{Results}

\subsubsection{Q1:}
\begin{figure}[t!]
  \centering
  \includegraphics[width=\linewidth]{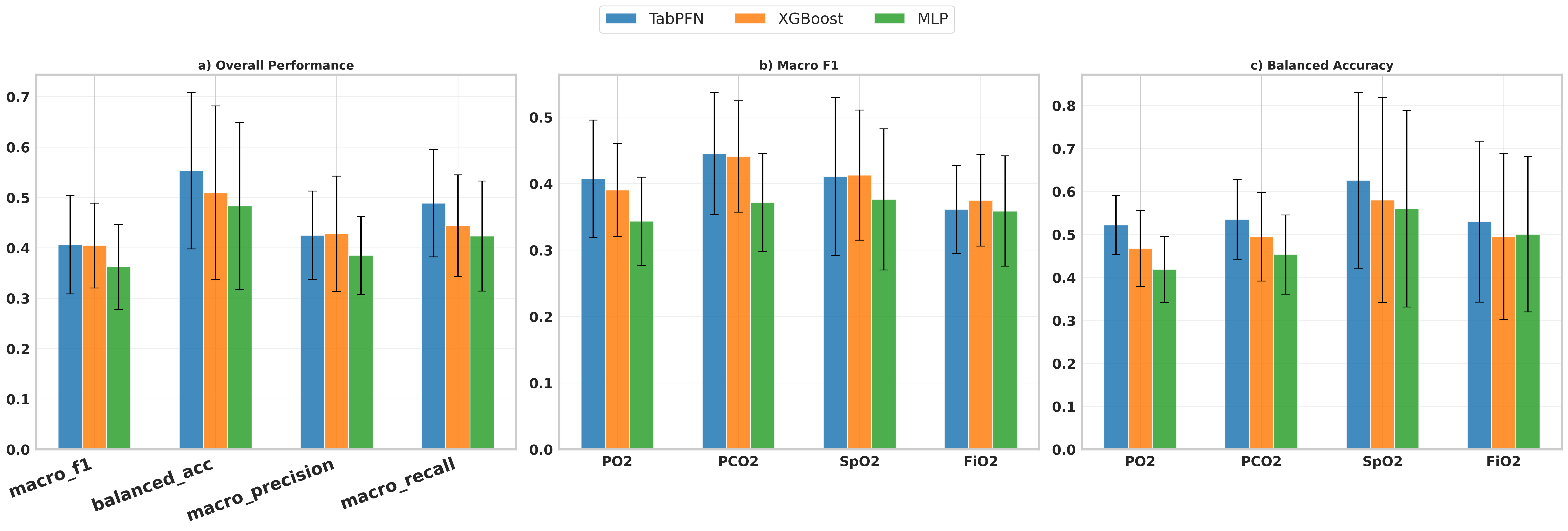}
  \caption{Overall performance aggregated across actions (mean $\pm$ std across LOO folds) and Per-knob F1 and accuracy (mean $\pm$ std across LOO folds) comparing TabPFN, XGBoost, and MLP}
  \label{fig:combined}
\end{figure}

TabPFN achieves the highest mean balanced accuracy and macro-F1 across most actionable features as shown in Figure~\ref{fig:combined}b and c. In particular, TabPFN demonstrates superior balanced accuracy across all targets, indicating improved robustness to class imbalance and generalization to held-out (unseen) patients. These improvements are also reflected in the aggregated overall metrics (Figure~\ref{fig:combined}a), supporting the claim that using the prior structure encoded in TabPFN yields more generalizable ECMO policies.



\subsubsection{Q2:}
\begin{wrapfigure}[13]{r}{5cm}
  \centering
  \includegraphics[width=4.5cm]{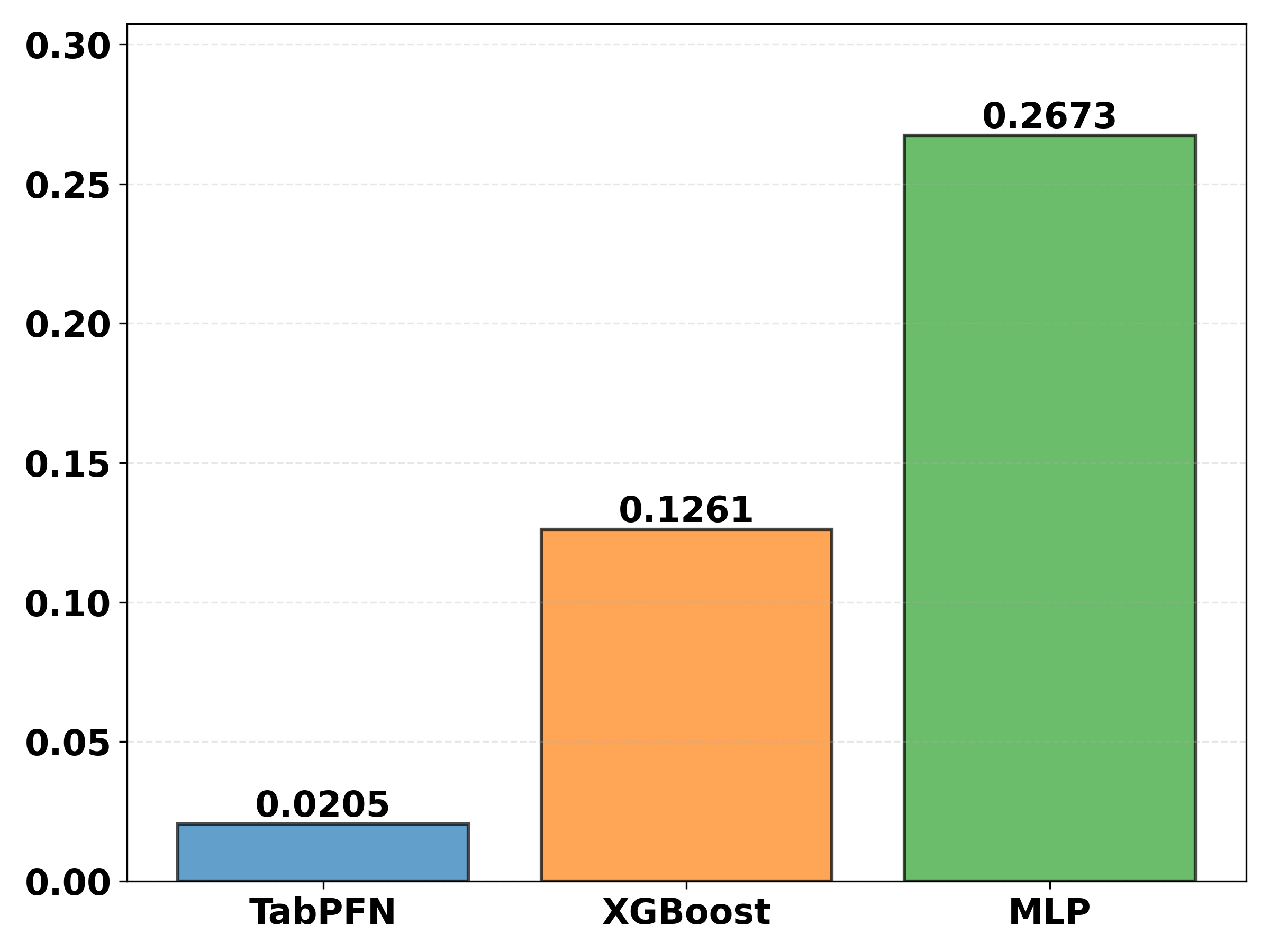}
  \caption{ECE for each model. Lower values indicate better alignment between predicted confidence and accuracy.}
  \label{fig:calibrated}
\end{wrapfigure}
To test policy calibration, we compute Expected Calibration Error (ECE) for each model. As shown in Figure~\ref{fig:calibrated}, TabPFN achieves substantially lower ECE compared to the baselines. This indicates that TabPFN's predictions are more calibrated and provide reliable confidence estimates.

\subsubsection{Q3:}

To find where model predictions diverge from clinician actions, we first define a per-sample disagreement target
$d(s) = \lvert y_{\text{c}} - \hat{p}_{\text{model}} \rvert$, where $y_{\text{c}}\in\{0,1\}$ is the clinician action label and $\hat{p}_{\text{model}}$ is the model-predicted probability for that action. Then, for each model and actionable feature, we train a shallow decision tree regressor (max depth 3) to predict $d(s)$ from the state features. 

By analyzing features in this way, we can identify which features are most often involved in model errors, implying they are misleading the model. For example, in TabPFN
, for the FiO$_2$ adjustment decision, TabPFN disagreements concentrated in a state with moderate FiO$_2$ and small recent changes:
\begin{align*}
\textbf{FiO}_2\ :\quad {FiO2} > 36\%, \quad -8\% < \Delta {FiO2} \le 13\%
\quad (n{=}986,\ d(s){=}0.11).
\end{align*} 

\begin{wraptable}[14]{r}{7cm}
\vspace{-0.8cm}
\centering
\caption{Most frequent split features appearing in disagreement trees, reported as the number of times each feature was selected across all knobs for each model.}
\label{tab:disagreement_split_features}
\small
\begin{tabular}{lccc}
\toprule
\textbf{Feature} & \textbf{MLP} & \textbf{TabPFN} & \textbf{XGBoost} \\
\midrule
\texttt{FiO2} & 2 & 5 & 5 \\
\texttt{SpO2} & 1 & 5 & 3 \\
\texttt{PH} & 2 & 3 & 2 \\
\texttt{PO2} & 3 & 2 & 1 \\
\texttt{PCO2} & 2 & 3 & 1 \\
\texttt{Lactate} & 3 & 1 & 1 \\
\bottomrule
\end{tabular}
\end{wraptable}



To summarize which signals most often defined disagreement regions, we additionally examined the most frequent features appearing in these regression-tree split rules (Table~\ref{tab:disagreement_split_features}). Across models, FiO$_2$ was the most common split feature, indicating that oxygen concentration consistently forms disagreement regions. TabPFN splits were dominated by FiO$_2$ and SpO$2$, whereas MLP disagreement was more frequently associated with PO$2$ and Lactate. This suggests that the models tend to deviate from clinician actions in different regions of state-space, motivating a further analysis of model-specific decision structure in future work.



\vspace{-1em}
\section{Conclusion}
We considered the problem of learning to act from pediatric ECMO trajectories for clinical decision support. This setting is hard because interventions are not explicitly recorded in the data, the action space is large and highly imbalanced, and patient-level data is limited. To address these challenges, we framed ECMO decision-making as offline imitation learning from observational trajectories. We trained a multi-head policy that predicts each knob’s action via a hierarchical two-stage pipeline (change detection followed by direction prediction), and compared classical tabular baselines (XGBoost and MLP) against TabPFN, a transformer-based prior-data fitted network. We find that TabPFN achieves the strongest generalization to held-out patients, with higher balanced accuracy and macro-F1 than XGBoost and MLP across knobs. TabPFN also produces better-calibrated policies, with lower expected calibration error. These results indicate that leveraging the prior structure encoded in a tabular foundation model can yield clinician-behavior policies that are both more generalizable and more calibrated in a data-scarce, high-stakes ICU setting. We additionally analyzed regions with high disagreement between model predictions and expert actions, and found that some baselines rely on different subsets of state features, suggesting qualitatively different feature-action mapping. Future work should (1) aggregate disagreement at the trajectory level and test whether model deviations correlate with neurological injury risk, (2) incorporate reward signals to learn better policies, and (3) perform a systematic analysis on the high-disagreement extracted rules and use them to elicit expert feedback for policy refinement.


%
%
%
\vspace{-1em}
\footnotesize
\bibliographystyle{splncs04}
\bibliography{refs}

\end{document}